# The Ethics of AI Ethics
## An Evaluation of Guidelines

Dr. Thilo Hagendorff

University of Tuebingen
International Center for Ethics in the Sciences and Humanities
thilo.hagendorff@uni-tuebingen.de

**Abstract** - Current advances in research, development and application of artificial intelligence (AI) systems have yielded a far-reaching discourse on AI ethics. In consequence, a number of ethics guidelines have been released in recent years. These guidelines comprise normative principles and recommendations aimed to harness the "disruptive" potentials of new AI technologies. Designed as a comprehensive evaluation, this paper analyzes and compares these guidelines highlighting overlaps but also omissions. As a result, I give a detailed overview of the field of AI ethics. Finally, I also examine to what extent the respective ethical principles and values are implemented in the practice of research, development and application of AI systems – and how the effectiveness in the demands of AI ethics can be improved.

## 1 Introduction

The current AI boom is accompanied by constant calls for applied ethics, which are meant to harness the "disruptive" potentials of new AI technologies. As a result, a whole body of ethical guidelines has been developed in recent years collecting principles, which technology developers should adhere to as far as possible. However, the critical question arises: Do those ethical guidelines have an actual impact on human decision-making in the field of AI and machine learning? The short answer is: No, most often not. This paper analyzes 21 of the major AI ethics guidelines and issues recommendations on how to overcome the relative ineffectiveness of these guidelines.

AI ethics – or ethics in general – lacks mechanisms to reinforce its own normative claims. Of course, the enforcement of ethical principles may involve reputational losses in the case of misconduct, or restrictions on memberships in certain professional bodies. Yet altogether, these mechanisms are rather weak and pose no eminent threat. Researchers, politicians, consultants, managers and activists have to deal with this essential weakness of ethics. However, it is also a reason why ethics is so appealing to many AI companies and institutions. When companies or research institutes formulate their own ethical guidelines, regularly incorporate ethical considerations into their public relations work, or adopt ethically motivated "self-commitments", efforts to create a truly binding legal framework are continuously discouraged. Ethics guidelines of the AI industry serve to suggest to legislators that internal self-governance in science and industry is sufficient, and that no specific laws are necessary to mitigate possible technological risks and to eliminate scenarios of abuse (Calo 2017). And even when more concrete laws concerning AI systems are demanded, as recently done by Google (Google 2019), these demands remain relatively vague and superficial.

Science- or industry-led ethics guidelines, as well as other concepts of self-governance, may serve to pretend that accountability can be devolved from state authorities and democratic institutions upon the respective sectors of science or industry. Moreover, ethics can also simply serve the purpose of calming critical voices from the public, while simultaneously the criticized practices are maintained within the organization. The association "Partnership on AI" (2018) which brings together companies such as Amazon, Apple, Baidu, Facebook, Google, IBM and Intel is exemplary in this context. Companies can highlight their membership in such associations whenever the notion of serious commitment to legal regulation of business activities needs to be stifled.

This prompts the question as to what extent ethical objectives are actually implemented and embedded in the development and application of AI, or whether merely good intentions are deployed. So far, some papers have been published on the subject of teaching ethics to data scientists (Garzcarek and Steuer 2019; Burton et al. 2017; Goldsmith and Burton 2017) but by and large very little to nothing has been written about the tangible implementation of ethical goals and values. In this paper, I address this question from a theoretical perspective. In a first step, 21 of the major guidelines of AI ethics will be analyzed and compared. I will also describe which issues they omit to mention. In a second step, I compare the principles formulated in the guidelines with the concrete practice of research and development of AI systems. In particular, I critically examine to what extent the principles have an effect. In a third and final step, I will work out ideas on how AI ethics can be transformed from a merely discursive phenomenon into concrete directions for action.

## 2 Guidelines in AI ethics

### 2.1 Method

Research in the field of AI ethics ranges from reflections on how ethical principles can be implemented in decision routines of autonomous machines (Anderson, M. and Anderson, S. Leigh 2015; Etzioni, A. and Etzioni, O. 2017; Yu, H. et al. 2018) over meta-studies about AI ethics (Vakkuri and Abrahamsson 2018; Prates, Avelar, and Lamb, Luis, C. 2018; Boddington 2017; Greene, Hoffman, and Stark 2019; Goldsmith and Burton 2017) or the empirical analysis on how trolley problems are solved (Awad et al. 2018) to reflections on specific problems (Eckersley

*Table 1: Overview of AI ethics guidelines and the different issues they cover*

2018) and comprehensive AI guidelines (The IEEE Global Initiative on Ethics of Autonomous and Intelligent Systems 2019). This paper deals with the latter issue. The list of ethics guidelines considered in this article therefore includes all such compilations that cover the field of AI ethics as comprehensively as possible. To the best of my knowledge, a few preprints and papers are currently available, which also deal with the comparison of different ethical guidelines (Zeng, Lu, and Huangfu 2018; Fjeld et al. 2019; Jobin, Ienca, and Vayena 2019).

The selection and compilation of 21 major ethical guidelines were based on a literature analysis. This selection was undertaken in two phases. In the first phase, I searched different databases, namely Google, Google Scholar, Web of Science, ACM Digital Library, arXiv, and SSRN for hits or articles on artificial intelligence and ethics. I took all documents no older than five years into account. In the second phase, I specifically selected those documents, which outline guidelines for an AI ethics or principles for an ethically justifiable application of AI systems. Documents that only refer to a national context – such as position papers of national interest groups (Smart Dubai 2018) or the report of the British House of Lords (Bakewell et al. 2018) – were excluded from the compilation. Nevertheless, I included the European Commission's "Ethics Guidelines for Trustworthy AI" (Pekka et al. 2018), the Obama administration's "Report on the Future of Artificial Intelligence" (Holdren et al. 2016), and the "Beijing AI Principles" (Beijing Academy of Artificial Intelligence 2019), which are backed by the Chinese Ministry of Science and Technology. I have included these three guidelines because they represent the three largest AI "superpowers". Furthermore, I included the "OECD Principles on AI" (Organisation for Economic Co-operation and Development 2019) due to their supranational character. Scientific papers or texts that fall into the category of AI ethics but focus on one or more specific aspects of the topic were not considered either. The same applies to guidelines or toolkits, which are not specifically about AI but rather about big data, algorithms or robotics (Anderson, D. et al. 2018; Anderson, M. and Anderson, S. Leigh 2011). I further excluded corporate policies, with the exception of the "Information Technology Industry AI Policy Principles" (2017), the principles of the "Partnership on AI" (2018), the IEEE first and second version of the document on "Ethically Aligned Design" (The IEEE Global Initiative on Ethics of Autonomous and Intelligent Systems 2016; The IEEE Global Initiative on Ethics of Autonomous and Intelligent Systems 2019), as well as the brief principle lists of Google (2018), Microsoft (2019), DeepMind (DeepMind), and IBM (2018) which have become well-known through media coverage. Furthermore, I included the "OpenAI Charta" (OpenAI 2018). Other large companies such as Facebook or Twitter have not yet published any systematic AI guidelines, but only isolated statements of good conduct. Paula Boddington's book on ethical guidelines (2017) funded by the Future of Life Institute was also not considered as it merely repeats the Asilomar principles (2017). The decisive factor for further selection of ethics guidelines was not the depth of detail of the individual document, but the discernible intention of a comprehensive mapping and categorization of normative claims with regard to the field of AI ethics. Thus, rather than listing and analyzing all AI ethics guidelines ever published within the scope of this paper, I have considered only those which have proven to be relevant in the international discourse. In table 1, I only inserted green markers if the corresponding issues were explicitly discussed in one or more paragraphs. Isolated mentions without further explanations were not considered, unless the analyzed guideline is so short that it consists entirely of brief mentions altogether.

## 2.2  Multiple entries

As shown in table 1, several issues are unsurprisingly recurring across various guidelines. Especially the aspects of *accountability*, *privacy* or *fairness* appear altogether in nearly 80% of all guidelines and seem to provide the minimal requirements for building and using an "ethically sound" AI system. What is striking here is the fact that the most frequently mentioned aspects are those for which technical fixes can be or have already been developed. Enormous technical efforts are undertaken to meet ethical targets in the fields of *accountability* and *explainable AI* (Mittelstadt, Russell, and Wachter 2019), *fairness* and *discrimination aware data mining* (Gebru et al. 2018), as well as *privacy* (Baron and Musolesi 2017). Many of those endeavors are unified under the FAT ML or XAI community (Veale and Binns 2017; Selbst et al. 2018). Several tech-companies already offer tools for *bias mitigation* and *fairness* in machine learning. In this context, Google, Microsoft and Facebook have issued the "AI Fairness 360" tool kit, the "What-If Tool", "Facets", "fairlern.py" and "Fairness Flow", respectively (Whittaker et al. 2018).

*Accountability*, *interpretability*, *privacy*, *justice*, but also other values such as *transparency*, *robustness* or *safety* are most easily operationalized mathematically and thus tend to be implemented in terms of technical solutions. With reference to the findings of psychologist Carol Gilligan, one could argue at this point that the way AI ethics is performed and structured constitutes a typical instantiation of a male-dominated justice ethics (Gilligan 1982). In the 1980s, Gilligan demonstrated in empirical studies that women do not, as men typically do, address moral problems primarily through a "calculating", "rational", "logic-oriented" ethics of justice, but rather interpret them within a wider framework of an "empathic", "emotion-oriented" ethics of care. In fact, no different from other parts of AI research, the discourse on AI ethics is also primarily shaped by men. My analysis of the distribution of female and male authors of the guidelines, as far as authors were indicated in the documents, showed that the proportion of women was only 37.1%. If the three AI Now Reports – which come from an organization, which is deliberately led by women – were left aside, the proportion of women would amount only to 31.3%. The proportion of women is lowest at 7.7% in the FAT ML community's guidelines which are focused predominantly on technical solutions (Diakopoulos et al.). Accordingly, the "male way" of thinking about ethical problems is reflected in almost all ethical guidelines by way of mentioning aspects such as *accountability*, *privacy* or *fairness*. In contrast, almost no guideline talks about AI in contexts of care, nurture, help, welfare, social responsibility or ecological networks. In AI ethics, technical artefacts are primarily seen as isolated entities that can be optimized by experts so as to find technical solutions for technical problems. What is often lacking is a consideration of the wider contexts and the comprehensive relationship networks in which technical systems are embedded. In accordance with that, it turns out that precisely the reports of AI Now (Whittaker et al. 2018; Campolo et al. 2017; Crawford et al. 2016), an organization primarily led by women, do not conceive AI applications in isolation, but within a larger network of social and ecological dependencies and relationships (Crawford and Joler 2018), corresponding most closely with the ideas and tenets of an ethics of care (Held 2013).

What are further insights from my analysis of the ethics guidelines, as summarized in table 1? On the one hand, it is noticeable that guidelines from industrial contexts name on average 9.1 distinctly separated ethical aspects, whereas the average for ethics codes from science is 10.5. The principles of Microsoft's AI ethics are the most brief and minimalistic (Microsoft Corporation 2019). The OpenAI Charta names only four points and is thus situated at the bottom of the list (OpenAI 2018). Conversely, the IEEE guideline contains the largest volume with more than 100.000 words (The IEEE Global Initiative on Ethics of Autonomous and Intelligent Systems 2019). Finally, yet importantly, it is noteworthy that almost all guidelines suggest that technical solutions exist for many of the problems described. Nevertheless, there are only two guidelines which contain genuinely technical explanations at all – albeit only very sparsely. The authors of the guideline on the "Malicious Use of AI" provide the most extensive commentary here (Brundage et al. 2018).

## 2.3 Omissions

Despite the fact that the guidelines contain various parallels and several recurring topics, what are issues the guidelines do not discuss at all or only very occasionally? The most striking aspect is that the danger of a *malevolent AGI* or *existential threats* is not discussed in any of the ethics guidelines, even though this is a thoroughly discussed topic in other contexts (Bostrom 2014; Tegmark 2017; Müller and Bostrom 2016; Omohundro 2014). This may be due to the fact that most of the guidelines are not written by research groups from philosophy or other speculative disciplines, but by researchers with a background directly in computer science or its application. In this context, it is noteworthy that the fear of the emergence of superintelligence is more frequently expressed by people who lack technical experience in the field of AI– one just has to think of people like Stephen Hawking, Elon Musk or Bill Gates – while "real" experts generally regard the idea of a strong AI as rather absurd (Calo 2017, 26). Perhaps the same holds true for the question of *machine consciousness* and the ethical problems associated with it (Lyons, S. 2018), as this topic is also omitted from all examined ethical guidelines. What is also striking is the fact that only the Montréal Declaration for Responsible Development of Artificial Intelligence (2018) addresses the aspect of democratic control, governance and political deliberation of AI systems. It is also the only guideline that explicitly prohibits imposing certain lifestyles or concepts of "good living" on people by AI systems, as it is for example demonstrated in the Chinese scoring system (Engelmann et al. 2019). It further criticizes the application of AI systems for the reduction of *social cohesion*, for example by isolating people in echo chambers (Flaxman, Goel, and Rao 2016). In addition, it is astonishing that hardly any guideline discusses the possibility for *political abuse of AI systems* in the context of automated propaganda, bots, fake news, deepfakes, micro targeting, election fraud, and the like. What is also largely absent from most guidelines is the issue of a *lack in diversity* within the AI community. This lack of diversity is prevailing in the field of artificial intelligence research and development, as well as in the workplace cultures shaping the technology industry. In the end, a relatively small group of predominantly white males determines how AI systems are designed, for what purposes they are optimized, what is attempted to realize technically, etc. The famous AI startup "nnaisense" run by Jürgen Schmidhuber, which aims at generating an artificial general intelligence (AGI), to name just one example, employs only two women – one scientist and one office

manager – in its team, but twenty-one men. Another matter, which is not covered at all or only very rarely mentioned in the guidelines, are aspects of *robot ethics*. As mentioned in the methods section, specific guidelines for robot ethics exist, most prominently represented by Asimov's three laws of robotics (Asimov 2004), but those guidelines were intentionally excluded from the analysis. Nonetheless, advances in AI research contribute, for instance, to increasingly anthropomorphized technical devices. The ethical question that arises in this context echoes Immanuel Kant's "brutalization argument" and states that the abuse of anthropomorphized agents – as, for example, is the case with language assistants (Brahnam 2006) – also promotes the likelihood of violent actions between people (Darling 2016). Apart from that, the examined ethics guidelines pay little attention to the rather popular *trolley problems* (Awad et al. 2018) and their alleged relation to ethical questions surrounding self-driving cars or other autonomous vehicles. In connection to this, no guideline deals in detail with the obvious question where systems of *algorithmic decision making* are superior or inferior, respectively, to human decision routines. And finally, virtually no guideline deals with the *"hidden" social and ecological costs* of AI systems. At several points in the guidelines, the importance of AI systems for approaching a sustainable society is emphasized (Rolnick et al. 2019). However, it is omitted that producer and consumer practices in the context of AI technologies may in themselves contradict sustainability goals. Issues such as lithium mining, e-waste, the one-way use of rare earth minerals, energy consumption, low-wage "clickworkers" creating labels for data sets or doing content moderation are of relevance here (Crawford and Joler 2018; Irani 2016; Veglis 2014; Fang 2019; Casilli 2017). Although "clickwork" is a necessary prerequisite for the application of methods of supervised machine learning, it is associated with numerous social problems (Silberman et al. 2018), such as low wages, work conditions and psychological work consequences, which tend to be ignored by the AI community. Finally, yet importantly, not a single guideline raises the issue of *public-private partnerships* and *industry-funded research* in the field of AI. Despite the massive lack of transparency regarding the allocation of research funds, it is no secret that large parts of university AI research are financed by corporate partners. In light of this, it remains questionable to what extent the ideal of freedom of research can be upheld – or whether there will be a gradual "buyout" of research institutes.

# 3 AI in practice
## 3.1 Business versus ethics

The close link between business and science is not only revealed by the fact that all of the major AI conferences are sponsored by industry partners. The link between business and science is also well illustrated by the AI Index 2018 (Shoham et al. 2018). Statistics show that, for example, the number of corporate-affiliated AI papers has grown significantly in recent years. Furthermore, there is a huge growth in the number of active AI startups, each supported by huge amounts of annual funding from Venture Capital firms. Tens of thousands of AI-related patents are registered each year. Different industries are incorporating AI applications in a broad variety of fields, ranging from manufacturing, supply-chain management, and service development, to marketing and risk assessment. All in all, the global AI market comprises more than seven billion dollars (Wiggers 2019).

A critical look at this global AI market and the use of AI systems in the economy and other social systems sheds light primarily on unwanted side effects of the use of AI, as well as on directly malevolent contexts of use. These occur in various areas (Pistono and Yampolskiy 2016; Amodei et al. 2017). Leading, of course, is the military use of AI in cyber warfare or regarding weaponized unmanned vehicles or drones (Ernest and Carroll 2016; Anderson, K. and Waxman 2013). According to media reports, the US government alone intends to invest two billion dollars in military AI projects over the next five years (Fryer-Biggs 2018). Moreover, governments can use AI applications for automated propaganda and disinformation campaigns (Lazer et al. 2018), social control (Engelmann et al. 2019), surveillance (Helbing 2019), face recognition or sentiment analysis (Introna and Wood 2004), social sorting (Lyon 2003), or improved interrogation techniques (McAllister 2017). Notwithstanding the above, companies can cause massive job losses due to AI implementation (Frey and Osborne 2013), conduct unmonitored forms of AI experiments on society without informed consent (Kramer, Guillory, and Hancock 2014), suffer from data breaches (Schneier 2018), use unfair, biased algorithms (Eubanks 2018), provide unsafe AI products (Sitawarin et al. 2018), use trade secrets to disguise harmful or flawed AI functionalities (Whittaker et al. 2018), rush to integrate and put immature AI applications on the market and many more. Furthermore, criminal or black-hat hackers can use AI to tailor cyberattacks, steal information, attack IT infrastructures, rig elections, spread misinformation for example through deepfakes, use voice synthesis technologies for fraud or social engineering (Bendel 2017), or disclose personal traits that are actually secret or private via machine learning applications (Kosinski and Wang 2018; Kosinski, Stillwell, and Graepel 2013; Kosinski et al. 2015). All in all, only a very small number of papers is published about the misuse of AI systems, even though they impressively show what massive damage can be done with those systems (Brundage et al. 2018; King et al. 2019; O'Neil 2016).

## 3.2 AI race

While the United States currently has the largest number of start-ups, China claims to be the "world leader in AI" in 2030 (abacus 2018). This claim is supported by the sheer amount of data that China has at its disposal to train its own AI systems, as well as by the large label companies that take over the manual preparation of data sets for supervised machine learning (Yuan 2018). Conversely, China is seen to have a weakness vis-à-vis the USA in that the investments of the market leaders Baidu, Alibaba and Tencent are too application-oriented comprising areas such as autonomous driving, finance or home appliances, while important basic research on algorithm development, chip production or sensor technology is neglected (Hao 2019). The constant comparison between China, the USA and Europe renders the fear of being inferior to each other an essential motive for efforts in the research and development of artificial intelligence.

Another justification for competitive thinking is provided by the military context. If the own "team", framed in a nationalist way, does not keep pace, so the consideration, it will simply be overrun by the opposing "team" with superior AI military technology. In fact, potential risks emerge from the AI race narrative, as well as from an actual competitive race to develop AI systems for technological superiority (Cave and ÓhÉigeartaigh 2018). One risk of this rhetoric is that "impediments" in the form of ethical considerations will be eliminated completely from research, development and implementation. AI research is not framed as a cooperative global project, but as a fierce competition. This competition affects the actions of individuals and promotes a climate of recklessness, repression, and thinking in hierarchies, victory and defeat. The race for the best AI, whether a mere narrative or a harsh reality, reduces the likelihood of the establishment of technical precaution measures as well as of the development of benevolent AI systems, cooperation, and dialogue between research groups and companies. Thus, the AI race stands in stark contrast to the idea of developing an "AI4people" (Floridi et al. 2018). The same holds true for the idea of an "AI for Global Good", as was proposed at the 2017's ITU summit, or the large number of leading AI researchers who signed the open letter of the "Future of Life Institute", embracing the norm that AI should be used for prosocial purposes.

Despite the downsides, in less public discourses and in concrete practice, an AI race has long since established itself. Along with that development, in- and outgroup-thinking has intensified. Competitors are seen more or less as enemies or at least as threats against which one has to defend oneself. Ethics, on the other hand, in its considerations and theories always stresses the danger of an artificial differentiation between in- and outgroups (Derrida 1997). Constructed outgroups are subject to devaluation, are perceived de-individualized and in the worst case can become victims of violence simply because of their status as "others" (Mullen and Hu 1989; Vaes, Bain, and Bastian 2014). I argue that only by abandoning such thinking in in- and outgroups may the AI race be reframed into a global cooperation for beneficial and safe AI.

## 3.3 Ethics in practice

Do ethical guidelines bring about a change in individual decision-making regardless of the larger social context? In a recent controlled study, researchers critically reviewed the idea that ethical guidelines serve as a basis for ethical decision-making for software engineers (McNamara, Smith, and Murphy-Hill 2018). In brief, their main finding was that the effectiveness of guidelines or ethical codes is almost zero and that they do not change the behavior of professionals from the tech community. In the survey, 63 software engineering students and 105 professional software developers were scrutinized. They were presented with eleven software-related ethical decision scenarios, testing whether the influence of the ethics guideline of the Association for Computing Machinery (ACM) (Gotterbarn et al. 2018) in fact influences ethical decision-making in six vignettes, ranging from responsibility to report, user data collection, intellectual property, code quality, honesty to customer to time and personnel management. The results are disillusioning: "No statistically significant difference in the responses for any vignette were found across individuals who did and did not see the code of ethics, either for students or for professionals." (McNamara, Smith, and Murphy-Hill 2018, 4)

Irrespective of such considerations on the microsociological level, the relative ineffectiveness of ethics can also be explained at the macrosociological level. Countless companies are eager to monetize AI in a huge variety of applications. This strive for a profitable use of machine learning systems is not primarily framed by value- or principle-based ethics, but obviously by an economic logic. Engineers and developers are neither systematically educated about ethical issues, nor are they empowered, for example by organizational structures, to raise ethical concerns. In business contexts, speed is everything in many cases and skipping ethical considerations is equivalent to the path of least resistance. Thus, the practice of development, implementation and use of AI applications has very often little to do with the values and principles postulated by ethics. The German sociologist Ulrich Beck once stated that ethics nowadays "plays the role of a bicycle brake on an intercontinental airplane" (Beck 1988, 194). This metaphor proves to be particularly true in the context of AI, where huge sums of money are invested in the development and commercial utilization of systems based on machine learning (Rosenberg 2017), while ethical considerations are mainly used for public relations purposes (Boddington 2017, 56).

In their AI Now 2017 Report, Kate Crawford and her team state that ethics and forms of soft governance "face real challenges" (Campolo et al. 2017, 5). This is mainly due to the fact that ethics has no enforcement mechanisms reaching beyond a voluntary and non-binding cooperation between ethicists and individuals working in research and industry. So what happens is that AI research and development takes place in "closed-door industry settings", where "user consent, privacy and transparency are often overlooked in favor of frictionless functionality that supports profit-driven business models" (Campolo et al. 2017, 31 f.). Despite this dispensation of ethical principles, AI systems are used in areas of high societal significance such as health, police, mobility or education. Thus, in the AI Now Report 2018, it is repeated that the AI industry "urgently needs new approaches to governance", since, "internal governance structures at most technology companies are failing to ensure accountability for AI systems" (Whittaker et al. 2018, 4). Thus, ethics guidelines often fall into the category of a "'trust us' form of [non-binding] corporate self-governance" (Whittaker et al. 2018, 30) and people should "be wary of relying on companies to implement ethical practices voluntarily" (Whittaker et al. 2018, 32).

The tension between ethical principles and wider societal interests on the one hand, and research, industry, and business objectives on the other can be explained with recourse to sociological theories. Especially on the basis of system theory it can be shown that modern societies differ in their social systems, each working with their own codes and communication media (Luhmann 1984; Luhmann 1997; Luhmann 1988). Structural couplings can lead decisions in one social system to influence other social systems. Such couplings, however, are limited and do not change the overall autonomy of social systems. This autonomy, which must be understood as an exclusive, functionalist orientation towards the system's own codes is also manifested in the AI industry, business and science. All these systems have their own codes, their own target values, and their own types of economic or symbolic capital via which they are structured and based upon which decisions are made (Bourdieu 1984). Ethical intervention in those systems is only possible to a very limited extent (Hagendorff 2016). A certain hesitance exists towards every kind of intervention as long as these lie beyond the functional laws of the respective systems. Despite that, unethical behavior or unethical intentions are not solely caused by economic incentives. Rather, individual character traits like cognitive moral development, idealism, or job satisfaction play a role, let alone organizational environment characteristics like an egoistic work climate or (non-existent) mechanisms for the enforcement of ethical codes (Kish-Gephart, Harrison, and Treviño 2010). Nevertheless, many of these factors are heavily influenced by the overall economic system logic. Ethics is then, so to speak, "operationally effectless" (Luhmann 2008).

And yet, such system-theoretical considerations apply only on a macro level of observation and must not be generalized. Deviations from purely economic behavioral logics in the tech industry occur as well, for example when Google withdrew from the military project "Maven" after protests from employees (Statt 2018) or when people at Microsoft protested against the company's cooperation with Immigration and Customs Enforcement (ICE) (Lecher 2018). Nevertheless, it must also be kept in mind here that, in addition to genuine ethical motives, the significance of economically relevant reputation losses should not be underestimated. Hence, the protest against unethical AI projects can in turn be interpreted in an economic logic, too.

## 3.4 Loyalty to guidelines

As indicated in the previous sections, the practice of using AI systems is poor in terms of compliance with the principles set out in the various ethical guidelines. Great progress has been made in the areas of privacy, fairness or explainability. For example, many privacy-friendly techniques for the use of data sets and learning algorithms have been developed, using methods where AI systems' "sight" is "darkened" via cryptography, differential or stochastic privacy (Ekstrand, Joshaghani, and Mehrpouyan 2018; Baron and Musolesi 2017; Duchi, Jordan, and Wainwright 2013; Singla et al. 2014). Nevertheless, this contradicts the observation that AI has been making such massive progress for several years precisely because of the large amounts of (personal) data available. Those data are collected by privacy-invasive social media platforms, smartphone apps, as well as Internet of Things devices with its countless sensors. In the end, I would argue that the current AI boom coincides with the emergence of a post-privacy society. In many respects, however, this post-privacy society is also a black box society (Pasquale 2015), in which, despite technical and organizational efforts to improve explainability, transparency and accountability, massive zones of non-transparency remain, caused both by the sheer complexity of technological systems and by strategic organizational decisions.

For many of the issues mentioned in the guidelines, it is difficult to assess the extent to which efforts to meet the set objectives are successful or whether conflicting trends prevail. This is the case in the areas of safety and cybersecurity, the science-policy link, future of employment, public awareness about AI risks, or human oversight. In other areas, including the issue of hidden costs and sustainability, the protection of whistleblowers, diversity in the field of AI, the fostering of solidarity and social cohesion, the respect for human autonomy, the use of AI for the common good or the military AI arms race, it can certainly be stated that the ethical goals are being massively underachieved. One only has to think of the aspect of gender diversity: Even though ethical guidelines clearly demand its improvement, the state of affairs is that on average 80% of the professors at the world's leading universities such as Stanford, Oxford, Berkeley or the ETH are male (Shoham et al. 2018). Furthermore, men make up more than 70% of applicants for AI jobs in the U.S. (Shoham et al. 2018). Alternatively, one can take human autonomy: As repeatedly demanded in various ethical guidelines, people should not be treated as mere data subjects, but as individuals. In fact, however, countless examples show that computer decisions, regardless of their susceptibility to error, are attributed a strong authority which results in the ignorance of individual circumstances and fates (Eubanks 2018). Furthermore, countless companies strive for the opposite of human autonomy, employing more and more subtle techniques for manipulating user behavior via micro targeting, nudging, UX-design and so on (Fogg 2003; Matz et al. 2017). Another example is that of cohesion: Many of the major scandals of the last years would have been unthinkable without the use of AI. From echo chamber effects (Pariser 2011) to the use of propaganda bots (Howard and Kollanyi 2016), or the spread of fake-news (Vosoughi, Roy, and Aral 2018), AI always played a key role to the effect of diminishing social cohesion, fostering instead radicalization, the decline of reason in public discourse and social divides (Tufekci 2018; Brady et al. 2017).

# 4 Advances in AI ethics
## 4.1 Technical instructions

Given the relative lack of tangible impact of the normative objectives set out in the guidelines, the question arises as to how the guidelines could be improved to make them more effective. At first glance, the most obvious potential for improvement of the guidelines is probably to supplement them with more detailed technical explanations – if such explanations can be found. Ultimately, it is a major problem to deduce concrete technological implementations from the very abstract ethical values and principles. What does it mean to implement justice or transparency in AI-systems? What does a "human-centered" AI look like? How can human oversight be obtained? The list of questions could easily be continued.

The ethics guidelines examined refer exclusively to the term "AI". They never or very seldom use more specific terminology. However, "AI" is just a collective term for a wide range of technologies or an abstract large-scale phenomenon. The fact that not a single prominent ethical guideline goes into greater technical detail shows how deep the gap is between concrete contexts of research, development, and application on the one side, and ethical thinking on the other. Ethicists must partly be capable of grasping technical details with their intellectual framework. That means reflecting on the ways data are generated, recorded, curated, processed, disseminated, shared, and used (Bruin and Floridi 2017), on the ways of designing algorithms and code, respectively (Kitchin 2017; Kitchin and Dodge 2011), or on the ways training data sets are selected (Gebru et al. 2018). In order to analyze all this in sufficient depth, ethics has to partially transform to "microethics". This means that at certain points, a substantial change in the level of abstraction has to happen insofar as ethics aims to have a certain impact and influence in the technical disciplines and the practice of research and development of artificial intelligence (Morley et al. 2019). On the way from ethics to "microethics", a transformation from ethics to technology ethics, to machine ethics, to computer ethics, to information ethics, to data ethics has to take place. As long as ethicists refrain from doing so, they will remain visible in a general public, but not in professional communities.

A good example of such a microethical work which can be implemented easily and concretely in practice is the paper by Gebru et al. (2018). The researchers propose the introduction of standardized datasheets listing the properties of different training data sets, so that machine learning-practitioners can check to what extent certain data sets are best suitable for their purposes, what the original intention was when the data set was created, what data the data set is composed of, how the data was collected and pre-processed, etc. The paper by Gebru et al. makes it possible for practitioners to obtain a more informed decision on the selection of certain training data sets, so that supervised machine learning ultimately becomes fairer, and more transparent, and avoids cases of algorithmic discrimination (Buolamwini and Gebru 2018). Such work is, however, an exception.

In general, ethical guidelines postulate very broad, overarching principles which are then supposed to be implemented in a widely diversified set of scientific, technical and economic practices, and in sometimes geographically dispersed groups of researchers and developers with different priorities, tasks and fragmental responsibilities. Ethics thus operates at a maximum distance from the practices it actually seeks to govern. Of course, this does not remain unnoticed among technology developers. In consequence, the generality and superficiality of ethical guidelines in many cases not only prevents actors from bringing their own practice into line with them, but rather encourages the devolution of ethical responsibility to others.

## 4.2 Virtue ethics

Regardless of the fact that normative guidelines should be accompanied by in-depth technical instructions – as far as they can reasonably be identified –, the question still arises how the precarious situation regarding the application and fulfillment of AI ethics guidelines can be improved. To address this question, one needs to take a step back and look at ethical theories in general. In ethics, several major strands of theories were created and shaped by various philosophical traditions. Those theories range from deontological to contractualistic, utilitarian, or virtue ethical approaches (Kant 1827; Rawls 1975; Bentham 1838; Hursthouse 2001). In the following, two of these approaches – deontology and virtue ethics – will be selected to illustrate different approaches in AI ethics. The deontological approach is based on strict rules, duties or imperatives. The virtue ethics approach, on the other hand, is based on character dispositions, moral intuitions or virtues. In the light of these two approaches, the traditional type of AI ethics can be assigned to the deontological concept (Mittelstadt 2019). Ethics guidelines postulate a fixed set of universal principles and maxims which technology developers should adhere to (Ananny 2016). The virtue ethics approach, on the other hand, focuses more on "deeper-lying" structures and situation-specific deliberations, on addressing personality traits and behavioral dispositions on the part of technology developers (Leonelli 2016). Virtue ethics does not define codes of conduct but focusses on the individual level. The technologists or software engineers and their social context are the primary addressees of such an ethics (Ananny 2016), not technology itself.

I argue that the prevalent approach of deontological AI ethics should be augmented with an approach oriented towards virtue ethics aiming at values and character dispositions. Ethics is then no longer understood as a deontologically inspired tick-box exercise, but as a project of advancing personalities, changing attitudes, strengthen responsibilities and gaining courage to refrain from certain actions, which are deemed unethical. When following the path of virtue ethics, ethics as a scientific discipline must refrain from wanting to limit, control, or steer (Luke 1995). Very often, ethics or ethical guidelines are perceived as something whose purpose is to stop or prohibit activity, to hamper valuable research and economic endeavors (Boddington 2017, 8). I want to resign this negative notion of ethics. It should not be the objective of ethics to stifle activity, but to do the exact opposite, i.e. broadening the scope of action, uncovering blind spots, promoting autonomy and freedom, and fostering self-responsibility.

Regardless of which approach to AI ethics is ultimately preferred, a closer link between research communities from ethics and other sciences is urgently needed. Researchers, academics or authors with an education in the humanities have typically acquired other vocabularies than researchers, academics or authors from scientific and technical subjects. In addition to that, an interactional distance exists between the two groups. While ethicists are often unwilling to acquire technical knowledge, technicians are also not generally or, in fact, rarely keen to learn from ethical considerations (McNamara, Smith, and Murphy-Hill 2018). For AI ethics to succeed, ethicists need to leave the "gravity-free" space of philosophical argument tournaments. Similarly, the establishment of a general willingness to learn and an openness towards ethical considerations on the part of the technicians is necessary. This entails not to marginalize ethics, but to perceive its considerations as actual enrichment for one's own practice. Ultimately, these considerations should be used to motivate and justify individual autonomy, empathy, and knowledge gains, but also concerns, deviations, or objections, as well as the appreciation of indirect technology consequences, social and ecological justice, and mechanisms of responsibility diffusion.

Especially the subject of responsibility diffusion can only be circumvented when virtue ethics is adopted on a broad and collective level in communities of tech professionals. Simply every person involved in data science, data engineering and data economies related to applications of AI has to take at least some responsibility for the implications of their actions (Leonelli 2016). This is why researchers such as Floridi argue that every actor who is causally relevant for bringing about the collective consequence or impacts in question, has to be held accountable (Floridi 2016). Interestingly, Floridi uses the backpropagation method known from Deep Learning to describe the way in which responsibilities can be assigned, except that here backpropagation is used in networks of distributed responsibility. When working in groups, actions that are on first glance allegedly morally neutral can nevertheless have consequences or impacts – intended or non-intended – that are morally wrong. This means that practitioners from AI communities always need to discern the overarching, short- and long-term consequences of the technical artefacts they are building or maintaining, as well as to explore alternative ways of developing software or using data, including the option of completely refraining from carrying out particular tasks, which are considered unethical.

In addition to the endorsement of virtue ethics in tech communities, several institutional changes should take place. They include the adoption of legal framework conditions, the establishment of mechanisms for an independent auditing of technologies, the establishment of institutions for complaints, which also compensate for harms caused by AI systems, and the expansion of university curricula in particular through content from ethics of technology, media, and information (Floridi et al. 2018; Cowls and Floridi 2018; Eaton et al. 2017; Goldsmith and Burton 2017). So far, however, hardly any of these demands have been met.

# 5 Conclusion

Currently, AI ethics is failing in many cases. Ethics lacks a reinforcement mechanism. Deviations from the various codes of ethics have no consequences. And in cases where ethics is integrated into institutions, it mainly serves as a marketing strategy. Furthermore, empirical experiments show that reading ethics guidelines has no significant influence on the decision-making of software developers. In practice, AI ethics is often considered as extraneous, as surplus or some kind of "add-on" to technical concerns, as unbinding framework that is imposed from institutions "outside" of the technical community. Distributed responsibility in conjunction with a lack of knowledge about long-term or broader societal technological consequences causes software developers to lack a feeling of accountability or a view of the moral significance of their work. Especially economic incentives are easily overriding commitment to ethical principles and values. This implies that the purposes for which AI systems are developed and applied are not in accordance with societal values or fundamental rights such as beneficence, non-maleficence, justice, and explicability (Taddeo and Floridi 2018; Pekka et al. 2018).

Nevertheless, in several areas ethically motivated efforts are undertaken to improve AI systems. This is particularly the case in fields where technical "fixes" can be found for specific problems, such as accountability, privacy protection, anti-discrimination, safety, or explainability. However, there is also a wide range of ethical aspects that are significantly related to the research, development and application of AI systems, but are not mentioned in the guidelines. Those omissions range from aspects like the danger of a malevolent AGI, machine consciousness, the reduction of social cohesion by AI ranking and filtering systems on social networking sites, the political abuse of AI systems, a lack of diversity in the AI community, links to robot ethics, the dealing with trolley problems, the weighting between algorithmic or human decision routines, "hidden" social and ecological costs of AI, to the problem of public-private-partnerships and industry-funded research.

Checkbox guidelines must not be the only "instruments" of AI ethics. A transition is required from a more deontologically oriented, action-restricting ethic based on universal abidance of principles and rules, to a situation-sensitive ethical approach based on virtues and personality dispositions, knowledge expansions, responsible autonomy and freedom of action. Such an AI ethics does not seek to subsume as many cases as possible under individual principles in an overgeneralizing way, but behaves sensitively towards individual situations and specific technical assemblages. Further, AI ethics should not try to discipline moral actors to adhere to normative principles, but emancipate them from potential inabilities to act self-responsibly on the basis of comprehensive knowledge, as well as empathy in situations where morally relevant decisions have to be made.

These considerations have two consequences for AI ethics. On the one hand, a stronger focus on technological details of the various methods and technologies in the field of AI and machine learning is required. This should ultimately serve to close the gap between ethics and technical discourses. It is necessary to build tangible bridges between abstract values and technical implementations, as long as these bridges can be reasonably constructed. On the other hand, however, the consequence of the presented considerations is that AI ethics, conversely, turns away from the description of purely technological phenomena in order to focus more strongly on genuinely social and personality-related aspects. AI ethics then deals less with AI as such, than with ways of deviation or distancing oneself from problematic routines of action, with uncovering blind spots in knowledge, and of gaining individual self-responsibility. Future AI ethics faces the challenge of achieving this balancing act between the two approaches.

# Acknowledgements

This research was supported by the Cluster of Excellence "Machine Learning – New Perspectives for Science" funded by the Deutsche Forschungsgemeinschaft (DFG, German Research Foundation) under Germany's Excellence Strategy – Reference Number EXC 2064/1 – Project ID 390727645.

# References

abacus. 2018. "China Internet Report 2018." Accessed July 13, 2018. https://www.abacusnews.com/china-internet-report/china-internet-2018.pdf.

Abrassart, Christophe, Yoshua Bengio, Guillaume Chicoisne, Nathalie de Marcellis-Warin, Marc-Antoine Dilhac, Sébastien Gambs, Vincent Gautrais et al. 2018. *Montréal Declaration for Responsible Development of Artificial Intelligence*: 1–21.

Amodei, Dario, Chris Olah, Jacob Steinhardt, Paul Christiano, John Schulman, and Dan Mané. 2017. "Concrete Problems in AI Safety." *arXiv*, 1–29.

Ananny, Mike. 2016. "Toward an Ethics of Algorithms: Convening, Observation, Probability, and Timeliness." *Science, Technology, & Human Values* 41 (1): 93–117.

Anderson, David, Joy Bonaguro, Miriam McKinney, Andrew Nicklin, and Jane Wiseman. 2018. *Ethics & Algoritms Toolkit*. Accessed February 01, 2019. https://ethicstoolkit.ai/.

Anderson, Kenneth, and Matthew C. Waxman. 2013. "Law and Ethics for Autonomous Weapon Systems: Why a Ban Won't Work and How the Laws of War Can." *SSRN Journal*, 1–32.

Anderson, Michael, and Susan Leigh Anderson, eds. 2011. *Machine Ethics.* Cambridge, MA: Cambridge University Press.

Anderson, Michael, and Susan Leigh Anderson. 2015. *Towards Ensuring Ethical Behavior from Autonomous Systems: A Case-Supported Principle-Based Paradigm*. Artificial Intelligence and Ethics: Papers from the 2015 AAAI Workshop: 1–10.

Asimov, Isaac. 2004. *I, Robot.* New York: Random House LLC.

Awad, Edmond, Sohan Dsouza, Richard Kim, Jonathan Schulz, Joseph Henrich, Azim Shariff, Jean-François Bonnefon, and Iyad Rahwan. 2018. "The Moral Machine Experiment." *Nature* 563 (7729): 59–64. https://doi.org/10.1038/s41586-018-0637-6.

Bakewell, Joan Dawson, Timothy Francis Clement-Jones, Anthony Giddens, Rosalind Mary Grender, Clive Richard Hollick, Christopher Holmes, Peter Keith Levene et al. 2018. *AI in the UK: Ready, Willing and Able?* Select Committee on Artificial Intelligence: 1–183.

Baron, Benjamin, and Mirco Musolesi. 2017. "Interpretable Machine Learning for Privacy-Preserving Pervasive Systems." *arXiv*, 1–10.

Beck, Ulrich. 1988. *Gegengifte: Die organisierte Unverantwortlichkeit.* Frankfurt am Main: Suhrkamp.

Beijing Academy of Artificial Intelligence. 2019. *Beijing AI Principles*. Accessed June 18, 2019. https://www.baai.ac.cn/blog/beijing-ai-principles.

Bendel, Oliver. 2017. "The Synthetization of Human Voices." *AI & SOCIETY - Journal of Knowledge, Culture and Communication* 82: 737.

Bentham, Jeremy. 1838. *The Works of Jeremy Bentham*. With the assistance of J. Bowring. 11 vols. 1. Edinburgh: William Tait. Published under the Superintendence of his Executor.

Boddington, Paula. 2017. *Towards a Code of Ethics for Artificial Intelligence.* Cham: Springer.

Bostrom, Nick. 2014. *Superintelligence: Paths, Dangers, Strategies.* Oxford: Oxford University Press.

Bourdieu, Pierre. 1984. *Distinction: A Social Critique of the Judgement of Taste.* Cambridge, Massachusetts: Harvard University Press.

Brady, William J., Julian A. Wills, John T. Jost, Joshua A. Tucker, and Jay J. van Bavel. 2017. "Emotion Shapes the Diffusion of Moralized Content in Social Networks." *Proceedings of the National Academy of Sciences of the United States of America* 114 (28): 7313–18.

Brahnam, Sheryl. 2006. "Gendered Bots and Bot Abuse." In *Misuse and Abuse of Interactive Technologies*, edited by Antonella de Angeli, Sheryl Brahnam, Peter Wallis, and Peter Dix, 1–4. Montreal: ACM.

Bruin, Boudewijn de, and Luciano Floridi. 2017. "The Ethics of Cloud Computing." *Science and Engineering Ethics* 23 (1): 21–39.

Brundage, Miles, Shahar Avin, Jack Clark, Helen Toner, Peter Eckersley, Ben Garfinkel, Allan Dafoe et al. 2018. *The Malicious Use of Artificial Intelligence: Forecasting, Prevention, and Mitigation*: 1–101.

Buolamwini, Joy, and Timnit Gebru. 2018. "Gender Shades: Intersectional Accuracy Disparities in Commercial Gender Classification." In Sorelle and Wilson 2018, 1–15.

Burton, Emanuelle, Judy Goldsmith, Sven Koening, Benjamin Kuipers, Nicholas Mattei, and Toby Walsh. 2017. "Ethical Considerations in Artificial Intelligence Courses." *Artificial Intelligence Magazine* 38 (2): 22–36.

Calo, Ryan. 2017. "Artificial Intelligence Policy: A Primer and Roadmap." *SSRN Journal*, 1–28.

Campolo, Alex, Madelyn Sanfilippo, Meredith Whittaker, and Kate Crawford. 2017. "AI Now 2017 Report." Accessed October 02, 2018. https://assets.ctfassets.net/8wprhhvnpfc0/1A9c3ZTCZa2KEYM64Wsc2a/8636557c5fb14f2b74b2be64c3ce0c78/_AI_Now_Institute_2017_Report_.pdf.

Casilli, Antonio A. 2017. "Digital Labor Studies Go Global: Toward a Digital Decolonial Turn." *International Journal of Communication* 11: 1934–3954.

Cave, Steven, and Seán S. ÓhÉigeartaigh. 2018. *An AI Race for Strategic Advantage: Rhetoric and Risks*: 1–5.

Cowls, Josh, and Luciano Floridi. 2018. "Prolegomena to a White Paper on an Ethical Framework for a Good AI Society." *SSRN Journal*, 1–14.

Crawford, Kate, and Vladan Joler. 2018. "Anatomy of an AI System." Accessed February 06, 2019. https://anatomyof.ai/.

Crawford, Kate, Meredith Whittaker, Madeleine Clare Elish, Solon Barocas, Aaron Plasek, and Kadija Ferryman. 2016. *The AI Now Report: The Social and Economic Implications of Artificial Intelligence Technologies in the Near-Term*.

Cutler, Adam, Milena Pribić, and Lawrence Humphrey. 2018. *Everyday Ethics for Artificial Intelligence: A Practical Guide for Designers & Developers*. Accessed February 04, 2019. https://www.ibm.com/watson/assets/duo/pdf/everydayethics.pdf: 1–18.

Darling, Kate. 2016. "Extending Legal Protection to Social Robots: The Effect of Anthropomorphism, Empathy, and Violent Behavior Towards Robotic Objects." In *Robot Law*, edited by Ryan Calo, A. M. Froomkin, and Ian Kerr, 213–34. Cheltenham, UK: Edward Elgar.

DeepMind. "DeepMind Ethics & Society Principles." Accessed July 17, 2019. https://deepmind.com/applied/deepmind-ethics-society/principles/.

Derrida, Jacques. 1997. *Of Grammatology.* Baltimore: Johns Hopkins Univ. Press.

Diakopoulos, Nick, Sorelle A. Friedler, Marcelo Arenas, Solon Barocas, Michael Hay, Bill Howe, Hosagrahar Visvesvaraya Jagadish et al. "Principles for Accountable Algorithms and a Social Impact Statement for Algorithms." Accessed July 31, 2019. https://www.fatml.org/resources/principles-for-accountable-algorithms.

Duchi, John C., Michael I. Jordan, and Martin J. Wainwright. 2013. "Privacy Aware Learning." *arXiv*, 1–60.

Eaton, Eric, Sven Koenig, Claudia Schulz, Francesco Maurelli, John Lee, Joshua Eckroth, Mark Crowley et al. 2017. "Blue Sky Ideas in Artificial Intelligence Education from the EAAI 2017 New and Future AI Educator Program." *arXiv*, 1–5.

Eckersley, Peter. 2018. "Impossibility and Uncertainty Theorems in AI Value Alignment or Why Your AGI Should Not Have a Utility Function." *arXiv*, 1–13.

Ekstrand, Michael D., Rezvan Joshaghani, and Hoda Mehrpouyan. 2018. "Privacy for All: Ensuring Fair and Equitable Privacy Protections." In Sorelle and Wilson 2018, 1–13.

Engelmann, Severin, Mo Chen, Felix Fischer, Ching-yu Kao, and Jens Grossklags. 2019. "Clear Sanctions, Vague Rewards: How China's Social Credit System Currently Defines "Good" and "Bad" Behavior." *Proceedings of the Conference on Fairness, Accountability, and Transparency - FAT\* '19*, 69–78.

Ernest, Nicholas, and David Carroll. 2016. "Genetic Fuzzy Based Artificial Intelligence for Unmanned Combat Aerial Vehicle Control in Simulated Air Combat Missions." *J Def Manag* 06 (01). https://doi.org/10.4172/2167-0374.1000144.

Etzioni, Amitai, and Oren Etzioni. 2017. "Incorporating Ethics into Artificial Intelligence." *J Ethics* 21 (4): 403–18.

Eubanks, Virginia. 2018. *Automating Inequality: How High-Tech Tools Profile, Police, and Punish the Poor.* New York: St. Marting's Press.

Fang, Lee. 2019. "Google Hired Gig Economy Workers to Improve Artificial Intelligence in Controversial Drone-Targeting Project." Accessed February 13, 2019. https://theintercept.com/2019/02/04/google-ai-project-maven-figure-eight/.

Fjeld, Jessica, Hannah Hilligoss, Nele Achten, Maia Levy Daniel, Joshua Feldman, and Sally Kagay. 2019. "Principled Artificial Intelligence: A Map of Ethical and Rights-Based Approaches." Accessed July 17, 2019. https://ai-hr.cyber.harvard.edu/primp-viz.html.

Flaxman, Seth, Sharad Goel, and Justin M. Rao. 2016. "Filter Bubbles, Echo Chambers, and Online News Consumption." *PUBOPQ* 80 (S1): 298–320.

Floridi, Luciano. 2016. "Faultless Responsibility: On the Nature and Allocation of Moral Responsibility for Distributed Moral Actions." *Philosophical transactions. Series A, Mathematical, physical, and engineering sciences* 374 (2083).

Floridi, Luciano, Josh Cowls, Monica Beltrametti, Raja Chatila, Patrice Chazerand, Virginia Dignum, Christoph Luetge et al. 2018. "AI4People - an Ethical Framework for a Good AI Society: Opportunities, Risks, Principles, and Recommendations." *Minds and Machines* 28 (4): 689–707.

Fogg, B. J. 2003. *Persuasive Technology: Using Computers to Change What We Think and Do.* San Francisco, California: Morgan Kaufmann Publishers.

Frey, Carl Benedikt, and Michael A. Osborne. 2013. *The Future of Employment: How Susceptible Are Jobs to Computerisation:* Oxford Martin Programme on Technology and Employment: 1–78.

Fryer-Biggs, Zachary. 2018. "The Pentagon Plans to Spend $2 Billion to Put More Artificial Intelligence into Its Weaponry." Accessed January 25, 2019. https://www.theverge.com/2018/9/8/17833160/pentagon-darpa-artificial-intelligence-ai-investment.

Future of Life Institute. 2017. "Asilomar AI Principles." Accessed October 23, 2018. https://futureoflife.org/ai-principles/.

Garzcarek, Ursula, and Detlef Steuer. 2019. "Approaching Ethical Guidelines for Data Scientists." *arXiv*, 1–18.

Gebru, Timnit, Jamie Morgenstern, Briana Vecchione, Jennifer Wortman Vaughan, Hanna Wallach, Hal Daumeé, III, and Kate Crawford. 2018. "Datasheets for Datasets." *arXiv*, 1–17.

Gilligan, Carol. 1982. *In A Different Voice: Psychological Theory and Women's Development.* Cambridge, Massachusetts: Harvard University Press.

Goldsmith, Judy, and Emanuelle Burton. 2017. *Why Teaching Ethics to AI Practitioners Is Important. ACM SIGCAS Computers and Society*: 110–14.

Google. 2018. "Artificial Intelligence at Google: Our Principles." Accessed January 24, 2019. https://ai.google/principles/.

Google. 2019. *Perspectives on Issues in AI Governance*. Accessed February 11, 2019. https://ai.google/static/documents/perspectives-on-issues-in-ai-governance.pdf: 1–34.

Gotterbarn, Don, Bo Brinkman, Catherina Flick, Michael S. Kirkpatrick, Keith Miller, Kate Vazansky, and Marty J. Wolf. 2018. *ACM Code of Ethics and Professional Conduct: Affirming Our Obligation to Use Our Skills to Benefit Society*. Accessed February 01, 2019. https://www.acm.org/binaries/content/assets/about/acm-code-of-ethics-booklet.pdf: 1–28.

Greene, Daniel, Anna Lauren Hoffman, and Luke Stark. 2019. "Better, Nicer, Clearer, Fairer: A Critical Assessment of the Movement for Ethical Artificial Intelligence and Machine Learning." *Hawaii International Conference on System Sciences*, 1–10.

Hagendorff, Thilo. 2016. "Wirksamkeitssteigerungen Gesellschaftskritischer Diskurse." *Soziale Probleme. Zeitschrift für soziale Probleme und soziale Kontrolle* 27 (1): 1–16.

Hao, Karen. 2019. "Three Charts Show How China's AI Industry Is Propped up by Three Companies." Accessed January 25, 2019. https://www.technologyreview.com/s/612813/the-future-of-chinas-ai-industry-is-in-the-hands-of-just-three-companies/?utm_campaign=Artificial%2BIntelligence%2BWeekly&utm_medium=email&utm_source=Artificial_Intelligence_Weekly_95.

Helbing, Dirk, ed. 2019. *Towards Digital Enlightment: Essays on the Darf and Light Sides of the Digital Revolution.* Cham: Springer.

Held, Virginia. 2013. "Non-contractual Society: A Feminist View." *Canadian Journal of Philosophy* 17 (Supplementary Volume 13): 111–37.

Holdren, John P., Afua Bruce, Ed Felten, Terah Lyons, and Michael Garris. 2016. *Preparing for the Future of Artificial Intelligence.* Washington, D.C. 1–58.

Howard, Philip N., and Bence Kollanyi. 2016. "Bots, #StrongerIn, and #Brexit: Computational Propaganda During the UK-EU Referendum." *arXiv*, 1–6.

Hursthouse, Rosalind. 2001. *On Virtue Ethics.* Oxford: Oxford University Press.

Information Technology Industry Council. 2017. *ITI AI Policy Principles*. Accessed January 29, 2019.

https://www.itic.org/public-policy/ITIAIPolicyPrinciplesFINAL.pdf.

Introna, Lucas D., and David Wood. 2004. "Picturing Algorithmic Surveillance: The Politics of Facial Recognition Systems." *Surveillance & Society* 2 (2/3): 177–98.

Irani, Lilly. 2016. "The Hidden Faces of Automation." *XRDS* 23 (2): 34–37.

Jobin, Anna, Marcello Ienca, and Effy Vayena. 2019. "The Global Landscape of AI Ethics Guidelines." *Nature Machine Intelligence* 1 (9): 389–99.

Kant, Immanuel. 1827. *Kritik Der Praktischen Vernunft*. Leipzig: Hartknoch.

King, Thomas C., Nikita Aggarwal, Mariarosaria Taddeo, and Luciano Floridi. 2019. "Artificial Intelligence Crime: An Interdisciplinary Analysis of Foreseeable Threats and Solutions." *Science and Engineering Ethics*, 1–36.

Kish-Gephart, Jennifer J., David A. Harrison, and Linda Klebe Treviño. 2010. "Bad Apples, Bad Cases, and Bad Barrels: Meta-Analytic Evidence About Sources of Unethical Decisions at Work." *The Journal of applied psychology* 95 (1): 1–31.

Kitchin, Rob. 2017. "Thinking Critically About and Researching Algorithms." *Information, Communication & Society* 20 (1): 14–29.

Kitchin, Rob, and Martin Dodge. 2011. *Code/Space: Software and Everyday Life.* Cambridge, Massachusetts: The MIT Press.

Kosinski, Michal, Sandra C. Matz, Samuel D. Gosling, Vesselin Popov, and David Stillwell. 2015. "Facebook as a Research Tool for the Social Sciences: Opportunities, Challenges, Ethical Considerations, and Practical Guidelines." *American Psychologist* 70 (6): 543–56.

Kosinski, Michal, David Stillwell, and Thore Graepel. 2013. "Private Traits and Attributes Are Predictable from Digital Records of Human Behavior." *Proceedings of the National Academy of Sciences of the United States of America* 110 (15): 5802–5.

Kosinski, Michal, and Yilun Wang. 2018. "Deep Neural Networks Are More Accurate Than Humans at Detecting Sexual Orientation from Facial Images." *Journal of Personality and Social Psychology* 114 (2): 246–57.

Kramer, Adam D. I., Jamie E. Guillory, and Jeffrey T. Hancock. 2014. "Experimental Evidence of Massive-Scale Emotional Contagion Through Social Networks." *Proceedings of the National Academy of Sciences of the United States of America* 111 (24): 8788–90.

Lazer, David M. J., Matthew A. Baum, Yochai Benkler, Adam J. Berinsky, Kelly M. Greenhill, Filippo Menczer, Miriam J. Metzger et al. 2018. "The Science of Fake News." *Science* 359 (6380): 1094–96.

Lecher, Colin. 2018. "The Employee Letter Denouncing Microsoft's ICE Contract Now Has over 300 Signatures." Accessed February 11, 2019. https://www.theverge.com/2018/6/21/17488328/microsoft-ice-employees-signatures-protest.

Leonelli, Sabina. 2016. "Locating Ethics in Data Science: Responsibility and Accountability in Global and Distributed Knowledge Production Systems." *Philosophical transactions. Series A, Mathematical, physical, and engineering sciences* 374 (2083).

Luhmann, Niklas. 1984. *Soziale Systeme: Grundriß einer allgemeinen Theorie.* Frankfurt a.M: Suhrkamp.

Luhmann, Niklas. 1988. *Die Wirtschaft der Gesellschaft.* Frankfurt a.M: Suhrkamp.

Luhmann, Niklas. 1997. *Die Gesellschaft der Gesellschaft.* Frankfurt am Main: Suhrkamp.

Luhmann, Niklas. 2008. *Die Moral der Gesellschaft.* Frankfurt a.M: Suhrkamp.

Luke, Brian. 1995. "Taming Ourselves or Going Feral? Toward a Nonpatriarchal Metaethic of Animal Liberation." In *Animals & Women: Feminist Theoretical Explorations*, edited by Carol J. Adams and Josephine Donovan, 290–319. Durham: Duke University Press.

Lyon, David. 2003. "Surveillance as Social Sorting: Computer Codes and Mobile Bodies." In *Surveillance as Social Sorting: Privacy, Risk, and Digital Discrimination*, edited by David Lyon, 13–30. London: Routledge.

Lyons, Siobhan. 2018. *Death and the Machine.* Singapore: Palgrave Pivot.

Matz, Sandra C., Michal Kosinski, Gideon Nave, and David Stillwell. 2017. "Psychological Targeting as an Effective Approach to Digital Mass Persuasion." *Proceedings of the National Academy of Sciences of the United States of America*, 1–6.

McAllister, Amanda. 2017. "Stranger Than Science Fiction: The Rise of A.I. Interrogation in the Dawn of Autonomous Robots and the Need for an Additional Protocol to the U.N. Convention Against Torture." *Minnesota Law Review* 101: 2527–73.

McNamara, Andrew, Justin Smith, and Emerson Murphy-Hill. 2018. "Does ACM's Code of Ethics Change Ethical Decision Making in Software Development?" In *Proceedings of the 2018 26th ACM Joint Meeting on European Software Engineering Conference and Symposium on the Foundations of*

Software Engineering - ESEC/FSE 2018, edited by Gary T. Leavens, Alessandro Garcia, and Corina S. Păsăreanu, 1–7. New York,: ACM Press.

Microsoft Corporation. 2019. "Microsoft AI Principles." Accessed February 01, 2019. https://www.microsoft.com/en-us/ai/our-approach-to-ai.

Mittelstadt, Brent. 2019. "AI Ethics – Too Principled to Fail?" *SSRN Journal*, 1–15.

Mittelstadt, Brent, Chris Russell, and Sandra Wachter. 2019. "Explaining Explanations in AI." *Proceedings of the Conference on Fairness, Accountability, and Transparency - FAT* '19*, 1–10.

Morley, Jessica, Luciano Floridi, Libby Kinsey, and Anat Elhalal. 2019. "From What to How. An Overview of AI Ethics Tools, Methods and Research to Translate Principles into Practices." *arXiv*, 1–21.

Mullen, Brian, and Li-tze Hu. 1989. "Perceptions of Ingroup and Outgroup Variability: A Meta-Analytic Integration." *Basic and Applied Social Psychology* 10 (3): 233–52.

Müller, Vincent C., and Nick Bostrom. 2016. "Future Progress in Artificial Intelligence: A Survey of Expert Opinion." In *Fundamental Issues of Artificial Intelligence*, edited by Vincent C. Müller, 555–72. Cham: Springer International Publishing.

Omohundro, Steve. 2014. "Autonomous Technology and the Greater Human Good." *Journal of Experimental & Theoretical Artificial Intelligence* (ahead-of-print): 1–13.

O'Neil, Cathy. 2016. *Weapons of Math Destruction: How Big Data Increases Inequality and Threatens Democracy.* New York: Crown Publishers.

OpenAI. 2018. "OpenAI Charter." Accessed July 17, 2019. https://openai.com/charter/.

Organisation for Economic Co-operation and Development. 2019. *Recommendation of the Council on Artificial Intelligence*. Accessed June 18, 2019. https://legalinstruments.oecd.org/en/instruments/OECD-LEGAL-0449: 1–12.

Pariser, Eli. 2011. *The Filter Bubble: What the Internet Is Hiding from You.* New York: The Penguin Press.

Partnership on AI. 2018. "About Us." Accessed January 25, 2019. https://www.partnershiponai.org/about/.

Pasquale, Frank. 2015. *The Black Box Society: The Sectret Algorithms That Control Money and Information.* Cambridge, Massachusetts: Harvard University Press.

Pekka, Ala-Pietlä, Wilhelm Bauer, Urs Bergmann, Mária Bieliková, Cecilia Bonefeld-Dahl, Yann Bonnet, Bouarfa. Loubna et al. 2018. *The European Commission's High-Level Expert Group on Artificial Intelligence: Ethics Guidelines for Trustworthy AI*. Working Document for stakeholders' consultation. Brussels: 1–37.

Pistono, Federico, and Roman Yampolskiy. 2016. "Unethical Research: How to Create a Malevolent Artificial Intelligence." *arXiv*, 1-6.

Prates, Marcelo, Pedro Avelar, and Lamb, Luis, C. 2018. "On Quantifying and Understanding the Role of Ethics in AI Research: A Historical Account of Flagship Conferences and Journals." *arXiv*, 1–13.

Rawls, John. 1975. *Eine Theorie Der Gerechtigkeit.* Frankfurt am Main: Suhrkamp.

Rolnick, David, Priya L. Donti, Lynn H. Kaack, Kelly Kochanski, Alexandre Lacoste, Kris Sankaran, Andrew Slavin Ross et al. 2019. "Tackling Climate Change with Machine Learning." *arXiv*, 1–97.

Rosenberg, Scott. 2017. "Why AI Is Still Waiting for Its Ethics Transplant." Accessed January 16, 2018. https://www.wired.com/story/why-ai-is-still-waiting-for-its-ethics-transplant/.

Schneier, Bruce. 2018. *Click Here to Kill Everybody.* New York: W. W. Norton & Company.

Selbst, Andrew D., danah boyd, Sorelle A. Friedler, Suresh Venkatasubramanian, and Janet Vertesi. 2018. "Fairness and Abstraction in Sociotechnical Systems." *ACT Conference on Fairness, Accountability, and Transparency (FAT)* 1 (1): 1–17.

Shoham, Yoav, Raymond Perrault, Eric Brynjolfsson, Jack Clark, James Manyika, Juan Carlos Niebles, Terah Lyons, John Etchemendy, Barbara Grosz, and Zoe Bauer. 2018. *The AI Index 2018 Annual Report.* Stanford, Kalifornien: 1–94.

Silberman, M. Six, Bill Tomlinson, Rochelle LaPlante, Joel Ross, Lilly Irani, and Andrew Zaldivar. 2018. "Responsible Research with Crowds." *Commun. ACM* 61 (3): 39–41.

Singla, Adish, Eric Horvitz, Ece Kamar, and Ryen W. White. 2014. "Stochastic Privacy." *arXiv*, 1–10.

Sitawarin, Chawin, Arjun Nitin Bhagoji, Arsalan Mosenia, Mung Chiang, and Prateek Mittal. 2018. "DARTS: Deceiving Autonomous Cars with Toxic Signs." *arXiv*, 1–27.

Smart Dubai. 2018. *AI Ethics Principles & Guidelines*. Accessed February 01, 2019. https://smartdubai.ae/pdfviewer/web/viewer.html?file=https://smartdubai.ae/docs/default-

source/ai-principles-resources/ai-ethics.pdf?Status=Master&sfvrsn=d4184f8d_6.

Sorelle, A. Friedler, and Christo Wilson, eds. 2018. *Proceedings of Machine Learning Research.* 81st ed. PMLR.

Statt, Nick. 2018. "Google Reportedly Leaving Project Maven Military AI Program After 2019." Accessed February 11, 2019. https://www.theverge.com/2018/6/1/17418406/google-maven-drone-imagery-ai-contract-expire.

Taddeo, Mariarosaria, and Luciano Floridi. 2018. "How AI Can Be a Force for Good." *Science* 361 (6404): 751–52.

Tegmark, Alex. 2017. *Life 3.0: Being Human in the Age of Artificial Intelligence.* New York: Alfred A. Knopf.

The IEEE Global Initiative on Ethics of Autonomous and Intelligent Systems. 2016. *Ethically Aligned Design: A Vision for Prioritizing Human Well-Being with Artificial Intelligence and Autonomous Systems*: 1–138.

The IEEE Global Initiative on Ethics of Autonomous and Intelligent Systems. 2019. *Ethically Aligned Design: A Vision for Prioritizing Human Well-Being with Autonomous and Intelligent Systems*: 1–294.

Tufekci, Zeynep. 2018. "YouTube, the Great Radicalizer." Accessed March 19, 2018. https://www.nytimes.com/2018/03/10/opinion/sunday/youtube-politics-radical.html.

Vaes, Jeroen, Paul G. Bain, and Brock Bastian. 2014. "Embracing Humanity in the Face of Death: Why Do Existential Concerns Moderate Ingroup Humanization?" *The Journal of Social Psychology* 154 (6): 537–45.

Vakkuri, Ville, and Pekka Abrahamsson. 2018. "The Key Concepts of Ethics of Artificial Intelligence." *Proceedings of the 2018 IEEE International Conference on Engineering, Technology and Innovation*, 1–6.

Veale, Michael, and Reuben Binns. 2017. "Fairer Machine Learning in the Real World: Mitigating Discrimination Without Collecting Sensitive Data." *Big Data & Society* 4 (2): 1-17.

Veglis, Andreas. 2014. "Moderation Techniques for Social Media Content." In *Social Computing and Social Media*, edited by David Hutchison, Takeo Kanade, Josef Kittler, Jon M. Kleinberg, Alfred Kobsa, Friedemann Mattern, John C. Mitchell et al., 137–48. Cham: Springer International Publishing.

Vosoughi, Soroush, Deb Roy, and Sinan Aral. 2018. "The Spread of True and False News Online." *Science* 359 (6380): 1146–51.

Whittaker, Meredith, Kate Crawford, Roel Dobbe, Genevieve Fried, Elizabeth Kaziunas, Varoon Mathur, Sarah Myers West, Rashida Richardson, Jason Schultz, and Oscar Schwartz. 2018. *AI Now Report 2018*: 1–62.

Wiggers, Kyle. 2019. "CB Insights: Here Are the Top 100 AI Companies in the World." Accessed February 11, 2019. https://venturebeat.com/2019/02/06/cb-insights-here-are-the-top-100-ai-companies-in-the-world/.

Yu, Han, Zhiqi Shen, Chunyan Miao, Cyril Leung, Voctor R. Lesser, and Qiang Yang. 2018. "Building Ethics into Artificial Intelligence." *arXiv*, 1–8.

Yuan, Li. 2018. "How Cheap Labor Drives China's A.I. Ambitions." Accessed November 30, 2018. https://www.nytimes.com/2018/11/25/business/china-artificial-intelligence-labeling.html.

Zeng, Yi, Enmeng Lu, and Cunqing Huangfu. 2018. "Linking Artificial Intelligence Principles." *arXiv*, 1–4.